\pdfoutput=1
\documentclass[11pt]{article}

\usepackage{acl}

\usepackage{times}
\usepackage{latexsym}

\usepackage[T1]{fontenc}
\usepackage[utf8]{inputenc}
\usepackage{adjustbox}
\usepackage{url}
\usepackage{amssymb}
\usepackage{amsmath}
\usepackage{multirow}
\usepackage{graphicx} %package to manage images
\usepackage{graphics}
\usepackage{multirow}
\usepackage{float}
\usepackage{fancyvrb}
\usepackage{amssymb}
\usepackage{amsmath}
\usepackage{amsfonts}
\usepackage{amssymb}
\usepackage{array}
\usepackage{ragged2e}
\usepackage{tabu}
\usepackage{algorithmicx}
\usepackage{algorithm}
\usepackage{algpseudocode}
\usepackage{adjustbox}
\usepackage{booktabs}
\usepackage{adjustbox}
\usepackage{subcaption}
\usepackage{pifont}
\newcommand{\cmark}{\ding{51}}
\newcommand{\xmark}{\ding{55}}

\usepackage{hyperref}\hypersetup{linkcolor=red,anchorcolor=blue,citecolor=blue}

\usepackage{microtype}

\title{Entity6K: A Large Open-Domain Evaluation Dataset for \\ Real-World Entity Recognition}

\author{Jielin Qiu$^{1,2}$, ~William Han$^{2}$, ~ Winfred Wang$^{2}$, ~Zhengyuan Yang$^{1}$, ~Linjie Li$^{1}$,\\
\textbf{~Jianfeng Wang$^{1}$,  ~Christos Faloutsos$^{2}$, ~Lei Li$^{2}$, ~Lijuan Wang$^{1}$ }\\
  $^{1}$ Microsoft Azure AI, 
  ~$^{2}$Carnegie Mellon University
  \\
  \small{$\left\{\text{jielinq,leili,christos}\right\}$@cs.cmu.edu, $\left\{\text{zhengyang,Lindsey.Li,jianfw,lijuanw}\right\}$@microsoft.com}
  }

\begin{document}
\maketitle
\begin{abstract}
Open-domain real-world entity recognition is essential yet challenging, involving identifying various entities in diverse environments. The lack of a suitable evaluation dataset has been a major obstacle in this field due to the vast number of entities and the extensive human effort required for data curation.
We introduce \textbf{Entity6K}, a comprehensive dataset for real-world entity recognition, featuring 5,700 entities across 26 categories, each supported by 5 human-verified images with annotations. Entity6K offers a diverse range of entity names and categorizations, addressing a gap in existing datasets.
We conducted benchmarks with existing models on tasks like image captioning, object detection, zero-shot classification, and dense captioning to demonstrate Entity6K's effectiveness in evaluating models' entity recognition capabilities. We believe Entity6K will be a valuable resource for advancing accurate entity recognition in open-domain settings.

\end{abstract}

\section{Introduction}

Recognizing entities from images is inherently difficult due to several factors. First, the visual complexity and variability of real-world scenes pose challenges in accurately identifying and localizing entities of interest. Images can contain multiple entities, occlusions, variations in lighting conditions, and diverse object appearances, making it challenging to discern and differentiate entities. Second, the task's open-domain nature demands models that can generalize across a wide range of entities, including those not seen during training, requiring abstract representations of entity characteristics across different visual contexts.

%Despite these challenges, open-domain entity recognition from images offers significant value across various domains, including scene understanding, object detection, visual search, recommendation systems, and augmented reality, enhancing user experiences and providing valuable insights from visual data. Additionally, entity recognition from images contributes to the development of intelligent systems, including image captioning, visual question answering, and content generation.

To address open-domain entity recognition in images, researchers have developed methods using deep learning and transfer learning, leveraging large-scale pretrained models. However, the lack of a comprehensive evaluation dataset hinders the assessment of different models' performance.
Creating such a dataset is challenging due to the need for a vast, diverse, and constantly updated list of entities, as well as the significant manual effort required for data curation. Additionally, the absence of standardized evaluation benchmarks impedes progress and makes it difficult to compare different approaches effectively.

\begin{figure}[t]
  \centering
  \includegraphics[width=0.99\linewidth]{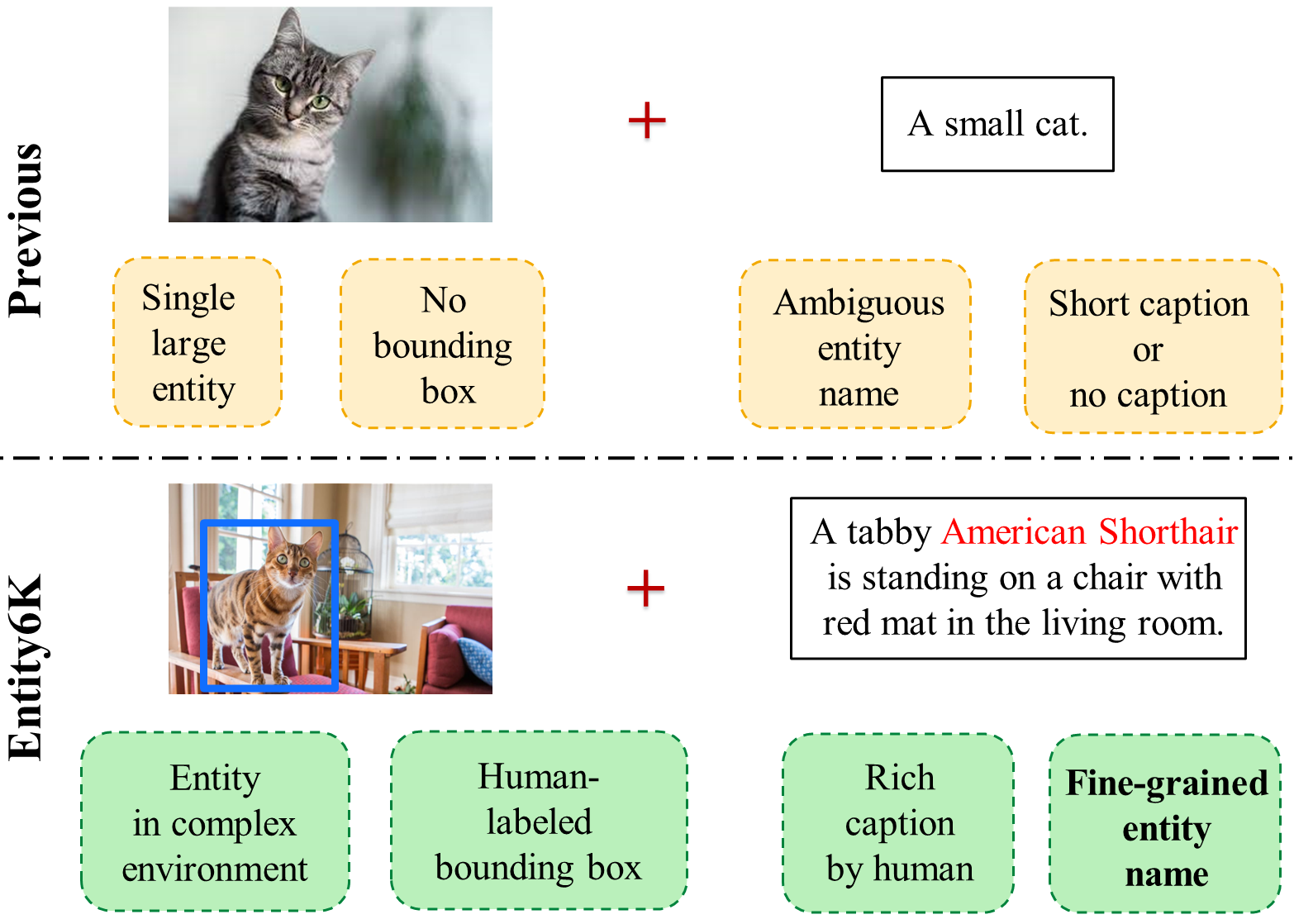}
  \caption{Comparison between \textbf{Entity6K} and existing datasets, where existing datasets may only contain a single large entity, ambiguous entity name, no bounding box, or short/no captions. However, our dataset contains entities in complex environments, with specific names, and human-labeled bounding boxes and captions.}
  \label{Fig:intro}
  \vspace{-15pt}
\end{figure}

Therefore, in this work, we introduce ``Entity6K," a large open-domain dataset specifically designed for the recognition of real-world entities. 
Our contributions can be summarized as follows:
\vspace{-5pt}
\begin{itemize}
    \item We introduced Entity6K, a comprehensive and diverse dataset containing 5,700 unique entities, providing a valuable resource for evaluating the entity recognition performance of various models.
    \vspace{-5pt}
   \item  
Each entity in the dataset is associated with five human-validated images and their corresponding annotations, resulting in a total of 28,500 images.
   \vspace{-5pt}
    \item We carried out benchmarking to assess pretrained models on tasks like image captioning, object detection, zero-shot classification, and dense captioning, highlighting their capabilities in recognizing real-world entities.
\end{itemize}

\vspace{-5pt}
\section{Related Work}

%\paragraph{Open-domain Entity Recognition} from images refers to the task of automatically identifying and extracting entities (objects, people, locations, etc.) from images without relying on any specific domain or prior knowledge.  There are few works in the open-domain entity recognition area. \cite{Hu2023OpendomainVE} presented the task of open-domain visual entity recognition, where a model needs to link an image to a Wikipedia entity with respect to a text query. However, their work needs a text query to retrieve the entity name in the Wikipedia entity name list. 

\paragraph{Open-domain Entity Recognition} in image processing involves automatically identifying various entities like objects, people, and locations in images. This task is challenging as it requires the system to work without domain-specific knowledge or predefined context.
\cite{Hu2023OpendomainVE} introduced a task where a model links an image to a Wikipedia entity using a text query. However, this method depends on a text query to retrieve the entity name from Wikipedia.

\paragraph{Zero-Shot Image Classification} involves recognizing unseen image classes, as explored in studies by \citet{Lampert2014AttributeBasedCF,Liu2019LargeScaleLR,Vinyals2016MatchingNF}. Due to its complexity, the few-shot learning approach, which utilizes limited training data, has been examined in works by \citet{Snell2017PrototypicalNF,Finn2017ModelAgnosticMF,Rusu2018MetaLearningWL,Ye2018FewShotLV}, focusing on developing effective models for this scenario.

\paragraph{Object Detection} techniques, like Faster R-CNN \citep{Ren2015FasterRT} and YOLO \citep{Redmon2015YouOL}, identify and localize objects in images, providing bounding boxes and class labels. F-VLM \citep{Kuo2022FVLMOO}, an open-vocabulary method, used Frozen Vision and Language Models. GLIP \citep{Li2021GroundedLP} merges object detection with phrase grounding for richer visual representations.\citet{Zhang2022GLIPv2UL} combines localization and Vision-Language pretraining for improved detection and segmentation. More related work is in Appendix~\ref{sec:appendix-more-related-work}.

%\paragraph{Image Segmentation} techniques can be employed to partition an image into different regions or segments corresponding to different entities. This approach can provide more fine-grained entity recognition by precisely delineating the boundaries of objects in an image. \cite{Kirillov2023SegmentA} lifted image segmentation into the era of foundation models. The model is designed and trained to be promptable, so it can transfer zero-shot to new image distributions and tasks. 

\section{Entity6K Dataset}

In this section, we explain the collection and annotation process of the Entity6K dataset. Detailed information is available in Appendix~\ref{sec:appendix-dataset-details}

\subsection{Data Acquisition}

\paragraph{Entity List}

To address our problem, we began by compiling a diverse set of entity names, covering a broad spectrum of real-world entities. We organized our selection into 26 distinct categories. Within each category, we used Wikipedia as a primary source to identify specific entity names. Our goal is to evaluate the system's ability to recognize precise entities accurately, so we focused on names with a high level of specificity. For example, we prefer names like ``German Shepherd" or ``Alaskan Malamute" over general terms like ``Dog." This approach sets our dataset apart from existing ones.

\begin{figure}[t]
  \centering
%  \fbox{\rule{0pt}{2in} \rule{0.9\linewidth}{0pt}}
  \includegraphics[width=0.99\linewidth]{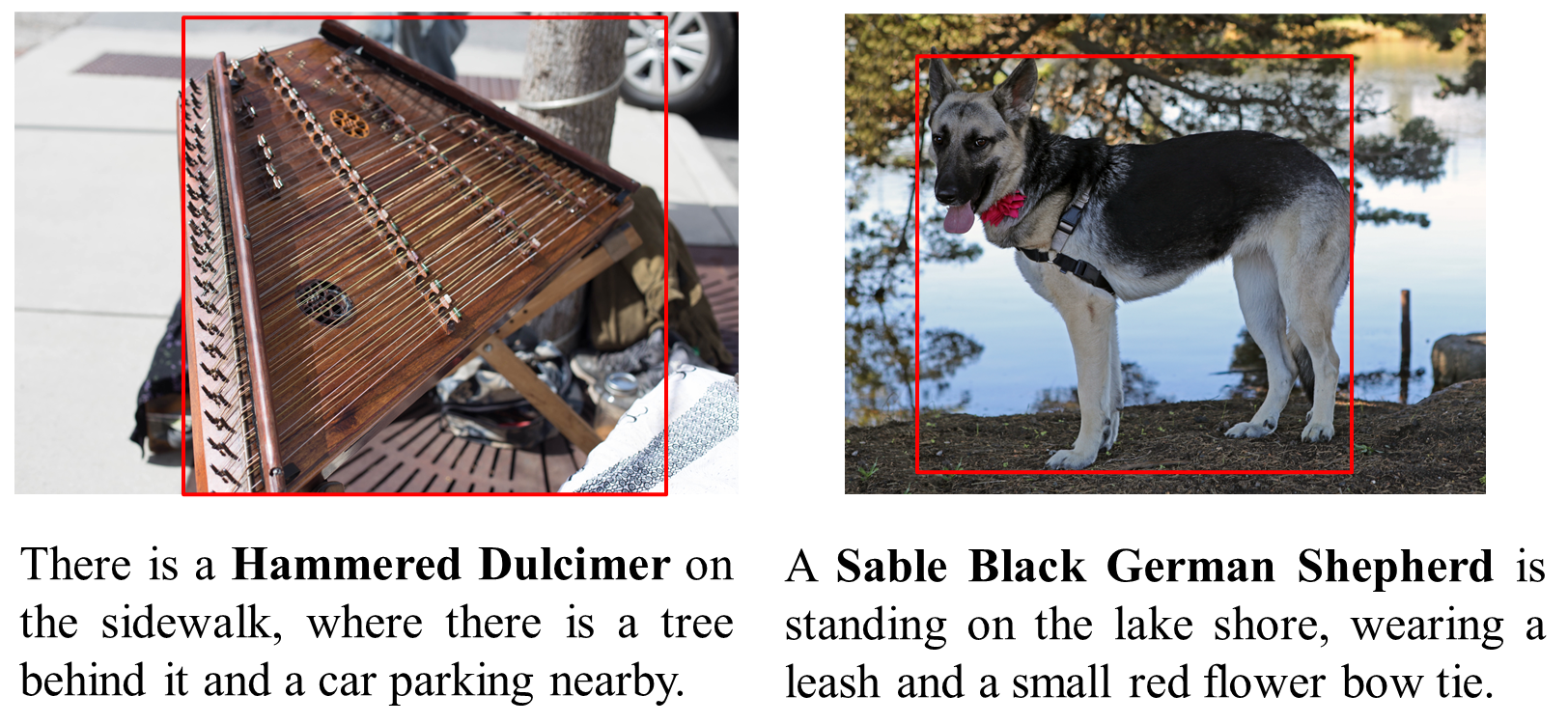}
  \vspace{-15pt}
  \caption{Examples of the collected data in the Entity6K dataset, where each image is associated with the entity region (bounding box) and the textual descriptions, \underline{centering on the specific entity}. }
  \label{Fig:example}
  \vspace{-5pt}
\end{figure}

\begin{table}[t]\small
\caption{Comparison with existing datasets, where HA is short for Human Annotations.}
\vspace{-5pt}
\centering
\begin{adjustbox}{width=0.99\linewidth}
\begin{tabular}{l|ccc}
\toprule
    Dataset & Entity & Categories & HA \\
\midrule
MSCOCO \citep{Lin2014MicrosoftCC} & 80 &  \xmark & \cmark  \\
ObjectNet \citep{Barbu2019ObjectNetAL} &313 & \xmark & \xmark  \\
SUN \citep{Xiao2010SUNDL} &397 & \xmark & \xmark  \\
Open Images \citep{Kuznetsova2018TheOI} &600 & \xmark & \cmark  \\
NoCaps \citep{Agrawal2019nocapsNO} &680 & \xmark & \cmark  \\
ImageNet \citep{Russakovsky2014ImageNetLS} & 1,000 & \xmark & \xmark \\ 
Entity6K (ours) & 5,700 & 26 & \cmark \\
\bottomrule
\end{tabular}
\label{table:compare_dataset}
\end{adjustbox}
\vspace{-5pt}
\end{table}

\vspace{-5pt}
\paragraph{Data Collection and Licenses}

After compiling a thorough and varied list of unique entities, ensuring there are no repetitions, the next step involves acquiring images. We accomplish this by utilizing the entity names as search queries on Flickr\footnote{https://www.flickr.com/photos/tags/dataset/}. It's important to note that these images have been generously shared on Flickr by their respective creators under licenses that include Creative Commons BY 2.0, Creative Commons BY-NC 2.0, Public Domain Mark, or Public Domain CC 1.0. These licenses all grant permission for unrestricted usage, redistribution, and modification, specifically for non-commercial purposes.

\vspace{-5pt}
\paragraph{Fidelity Control}
The dataset contains 28,500 high-quality images from Flickr, reflecting the diversity and biases of that database. Initially, we compiled 12,003 entity names across 26 categories, collecting ten images per entity with approved licenses. Using Amazon Mechanical Turk\footnote{https://www.mturk.com/}, we assessed image quality through two steps: (1) Three human judges verified the accuracy of each image in representing its entity, deleting any mismatches. (2) Entities with fewer than five accurate images were removed. For entities with more than five accurate images, five were randomly selected for the final dataset. After this quality control, we retained 5,700 entities, resulting in a retention rate of approximately 47.5\%. The detailed numbers of entities in each category before and after this process are shown in Table~\ref{table:fidelity_control} in Appendix~\ref{sec:appendix-dataset-details}.

\subsection{Human Annotation}

The dataset labeling process comprises two distinct stages with Amazon Mechanical Turk:

\paragraph{Bounding Box Annotation} 
In the initial phase, a single annotator is assigned to outline bounding boxes for each image. The annotator is given the corresponding entity name for the image and is responsible for marking the relevant region within that image. The objective is to establish a single bounding box for each image.

\paragraph{Textual Description Annotation} 
Following the completion of the initial bounding box marking phase by the first annotator, the second step involves five different annotators independently creating textual descriptions for each image. These annotators are given the entity name associated with each image to assist them in crafting their text captions. It's crucial to emphasize that all annotators are expected to provide comprehensive and detailed textual descriptions, encompassing as much relevant information as possible. For example, annotators are encouraged to write descriptions such as ``A cheerful boy, wearing a white helmet, is riding a vibrant green bicycle, while nearby, a young girl in a pink helmet is seated on a serene blue bicycle, sipping refreshing water" rather than simply stating ``Two people riding bikes."

\subsection{Statistics of the Dataset}

In Figure~\ref{Fig:statistics} in Appendix~\ref{sec:appendix-dataset-details}, we present the statistics of the collected Entity6K dataset. Furthermore, Table~\ref{table:compare_dataset} compares our dataset with existing datasets, which shows that our dataset contains an order of magnitude more entities than the existing datasets. Additionally, the entities are categorized and come with verified human annotations, rendering the proposed dataset a valuable resource for real-world entity recognition evaluations.

\section{Experimental Settings}

\subsection{Tasks}

We have chosen four tasks to construct our evaluation benchmark, which includes object detection, zero-shot image classification, image captioning, and dense captioning.

\subsection{Evaluation Metrics} According to different tasks, we select the corresponding standard metrics as the evaluation metrics. For object detection, we select Average Precision (AP) as the evaluation metric. For zero-shot image classification, we take the standard accuracy as the evaluation metric. For image captioning, we adopted the BLEU \citep{Papineni2002BleuAM}, ROUGE \citep{Lin2004ROUGEAP}, Meteor \citep{Banerjee2005METEORAA}, and BertScore \citep{Zhang2020BERTScoreET} as evaluation metrics.
For the dense captioning task, we take mean Average Precision (mAP) as the evaluation metric. Similar to object detection metric, dense captioning measures an mAP across a range of thresholds for both localization and description accuracy, following \cite{Johnson2015DenseCapFC}. For localization, it uses box IoU thresholds of .3, .4, .5, .6, .7. For language
description, a METEOR score \citep{Banerjee2005METEORAA} with thresholds of 0, .05, .1, .15, .2, .25 is used. The mAP is averaged by the APs across all pairwise of these two types of thresholds.

\subsection{Benchmark Models}

For different tasks, we selected different baseline models for the benchmark. Specifically, for object detection, GLIP~\citep{Li2021GroundedLP}, GRiT~\citep{Wu2022GRiTAG}, DINO~\citep{zhang2022dino}, and ViT-Adapter \citep{chen2022vitadapter}. For zero-shot image classification, we select CLIP \citep{Radford2021LearningTV}, ALIGN \citep{Jia2021ScalingUV}, and GPT-4 \citep{OpenAI2023GPT4TR}. 
For image captioning, we select BLIP~\citep{Li2022BLIPBL}, OFA~\citep{Wang2022UnifyingAT}, GIT~\citep{Wang2022GITAG}, and GRIT~\cite{Nguyen2022GRITFA} as baselines. For dense captioning, we adopt FCLN~\cite{Johnson2015DenseCapFC} and GRiT~\cite {Wu2022GRiTAG}. Details about the baseline models can be found in the Appendix.

\subsection{Experimental Settings}

In our evaluation of the performance of existing models, we adhered to the instructions provided by those models. Specifically, we utilized the pretrained weights directly without undergoing any training or fine-tuning processes.

\section{Experimental Results}

\subsection{General Insights}

In this section, we provide comparison results and discussions on each task.

\begin{table}[t]\small
\caption{Averaged Object Detection results.}
\vspace{-5pt}
\centering
\begin{adjustbox}{width=0.95\linewidth}
\begin{tabular}{l|rrr} 
\toprule
Method & AP & AP$_{50}$ & AP$_{75}$  \\ 
\midrule
GLIP~\citep{Li2021GroundedLP} & 8.90 & 12.54 & 0.04\\
DINO~\citep{zhang2022dino} &10.82 & 14.42 & 2.37\\
ViT-Adapter~\citep{chen2022vitadapter} & 11.83 & 16.77 & 6.90\\
GRiT~\citep{Wu2022GRiTAG}  &\textbf{14.41} & \textbf{23.30} & \textbf{7.89} \\
\bottomrule
\end{tabular}
\label{table:obj_detect_results}
\end{adjustbox}
\vspace{-5pt}
\end{table}

\vspace{-5pt}
\paragraph{Object Detection}

The Object Detection results are presented in Table~\ref{table:obj_detect_results}. According to the findings, GRiT outperforms all other baselines across all metrics.

\begin{table}[t]\small
\caption{Ave. Zero-shot Image Classification results.}
\vspace{-5pt}
\centering
\begin{adjustbox}{width=0.8\linewidth}
\begin{tabular}{l|c} 
\toprule
Method & Acc (\%) \\ 
\midrule
ALIGN \citep{Jia2021ScalingUV} & 34.66\\
CLIP-ViT-L \citep{Radford2021LearningTV} &   54.10\\
CLIP-ViT-H \citep{Radford2021LearningTV}  & 57.01 \\
GPT-4 \citep{OpenAI2023GPT4TR} & \textbf{69.25}  \\
\midrule
Human & 71.25 \\
\bottomrule
\end{tabular}
\label{table:classfication_results}
\end{adjustbox}
\vspace{-13pt}
\end{table}

\vspace{-5pt}
\paragraph{Zero-shot Image Classification}
The Zero-shot Image Classification results are outlined in Table 2. CLIP outperforms ALIGN, and CLIP with the ViT-H vision encoder shows better performance than CLIP with the ViT-L vision encoder, suggesting that a larger vision encoder can learn more effective visual representations. However, GPT-4 achieved the best performance compared to all the baselines, demonstrating its superior ability to recognize real-world entities.

\vspace{-5pt}
\paragraph{Image Captioning}
As shown in Table 3, various models show different performances across evaluation metrics. BLIP surpasses others in ROUGE-1, BLEU, and METEOR, while OFA outperforms BLIP in ROUGE-2, ROUGE-L, SPICE, and BertScore metrics.

\begin{table}[t]\small
\caption{Averaged Image Captioning results. }
\vspace{-5pt}
\centering
\begin{adjustbox}{width=0.99\linewidth}
\begin{tabular}{l|ccccc} 
\toprule
Methods    & ROUGE-L $\uparrow$ & BLEU $\uparrow$ & METEOR $\uparrow$ & SPICE $\uparrow$ & BertScore $\uparrow$  \\ 
\midrule
GRIT~\cite{Nguyen2022GRITFA} & 0.01 & 0.01 & 0.20 & 0.13 & 77.85 \\
GIT~\citep{Wang2022GITAG} & 9.92 & 0.40 & 4.37  & 1.27 & 81.34 \\
BLIP~\citep{Li2022BLIPBL} & 11.67 & \textbf{1.11} & \textbf{7.75}  & 1.74 & 84.52 \\
OFA~\citep{Wang2022UnifyingAT} & \textbf{12.02} & 0.92 & 6.89  & \textbf{3.27} & \textbf{84.63} \\
%GPT-4 \\
\bottomrule
\end{tabular}

\label{table:imgcap_results}
\end{adjustbox}
\vspace{-5pt}
\end{table}

\vspace{-5pt}
\paragraph{Dense Captioning }
In Table~\ref{table:densecap_results}, while GRiT outperforms FCLN, it's noteworthy that the results of both models are relatively low, indicating significant room for improvement in this area.

\begin{table}[t]\small
\caption{Averaged Dense Captioning results.}
\centering
\vspace{-5pt}
\begin{adjustbox}{width=0.6\linewidth}
\begin{tabular}{l|c} 
\toprule
Method & mAP  \\ 
\midrule
FCLN \citep{densecap} & 0.02\\
GRiT$_{\text{MAE}}$ \citep{Wu2022GRiTAG} & \textbf{2.12} \\
\midrule
Human & 20.12 \\
\bottomrule
\end{tabular}
\label{table:densecap_results}
\end{adjustbox}
\vspace{-10pt}
\end{table}

\subsection{Detailed Results for Each Category}

The detailed results for each category on each task are listed in Appendix~\ref{sec:appendix-more-results}. 
An important observation across these results is that the prevalence of a category in our dataset does not directly correlate to performance. For example, cars and birds comprise 2.3\% and 12.4\% of our dataset, respectively. However, in most results, the metrics for the birds category are often lower than the cars category. We assume this is due to each model being pretrained on different datasets. Overall, by observing the category-wise performances of all models for each task, we can conclude that none of the models can generalize well to the complex scenes and textual descriptions provided in our dataset, highlighting the complexity and challenge of our proposed dataset.

\subsection{Human evaluation}

To improve the model's performance assessment, we conducted human experiments for both the Zero-shot Image Classification and Dense Captioning tasks. We engaged three human judges from Amazon Mechanical Turk, including two males and one female. The results for each task were derived by averaging the scores provided by all three human judges and are detailed in Table~\ref{table:classfication_results} and Table~\ref{table:densecap_results}.
We can see that GPT-4 has achieved performance levels closely resembling human capabilities in the Zero-shot Image Classification task. However, it's worth noting that in the Dense Captioning task, both models' results fall significantly below human performance levels. This indicates a considerable scope for improvement in this specific domain.

\section{Conclusion}

In this study, we investigated the open-domain recognition capabilities of pretrained multimodal models. To aid this investigation, we introduced Entity6K, a large open-domain dataset designed for real-world entity recognition. With 5,700 diverse real-world entities across 26 distinct categories, this dataset is versatile and applicable to various tasks. We conducted evaluations of model performance across four tasks: image captioning, object detection, zero-shot image classification, and dense captioning. Our goal with these evaluations is to offer a valuable resource for assessing models' proficiency in recognizing open-domain real-world entities.

\clearpage
\section*{Limitations}

Although our proposed dataset tackles the shortcomings of current datasets, we foresee that there are still certain limitations that future research can potentially improve.
\begin{itemize}
    \item The dataset size has the potential to be expanded further. Although we initially compiled a substantial list of entities, our fidelity control process led to the removal of over half of the entity names due to insufficient images. To address this issue, future endeavors could explore additional resources beyond the Flickr database we utilized, with the aim of augmenting the dataset.
    \item Achieving data balance remains a challenge. Despite our efforts to create a diverse dataset, imbalances between different categories may persist. Future efforts could focus on balancing entities within each category while expanding the dataset. However, it's important to note that certain categories, like species of mammals, may inherently have limited entities, while others, such as celebrity names, could be significantly larger. This inherent nature might lead to persistent imbalances in the enlarged dataset.
    \item Insufficient baseline options, particularly in the context of dense captioning, pose a challenge. Currently, only two baselines with publicly available weights can be incorporated into this benchmark. It is anticipated that future research endeavors could expand the available baseline options as new work emerges, providing a more comprehensive selection for evaluation.
\end{itemize}

\section*{Data availability statement}

In this paper, we introduced Entity6K, a large open-domain evaluation dataset for real-world entity recognition. Entity6K contains 5,700 real-world entities with 26 main categories, where each entity is associated with five human-verified images and human annotations/captions. Our dataset will be made publicly available soon.

% Entries for the entire Anthology, followed by custom entries
\bibliography{egbib}

\begin{thebibliography}{43}
\expandafter\ifx\csname natexlab\endcsname\relax\def\natexlab#1{#1}\fi

\bibitem[{Agrawal et~al.(2019)Agrawal, Desai, Wang, Chen, Jain, Johnson, Batra,
  Parikh, Lee, and Anderson}]{Agrawal2019nocapsNO}
Harsh Agrawal, Karan Desai, Yufei Wang, Xinlei Chen, Rishabh Jain, Mark
  Johnson, Dhruv Batra, Devi Parikh, Stefan Lee, and Peter Anderson. 2019.
\newblock \href {https://api.semanticscholar.org/CorpusID:56517630} {nocaps:
  novel object captioning at scale}.
\newblock \emph{International Conference on Computer Vision}, pages 8947--8956.

\bibitem[{Banerjee and Lavie(2005)}]{Banerjee2005METEORAA}
Satanjeev Banerjee and Alon Lavie. 2005.
\newblock Meteor: An automatic metric for mt evaluation with improved
  correlation with human judgments.
\newblock In \emph{IEEvaluation@ACL}.

\bibitem[{Barbu et~al.(2019)Barbu, Mayo, Alverio, Luo, Wang, Gutfreund,
  Tenenbaum, and Katz}]{Barbu2019ObjectNetAL}
Andrei Barbu, David Mayo, Julian Alverio, William Luo, Christopher Wang, Dan
  Gutfreund, Joshua~B. Tenenbaum, and Boris Katz. 2019.
\newblock \href {https://api.semanticscholar.org/CorpusID:202777185}
  {Objectnet: A large-scale bias-controlled dataset for pushing the limits of
  object recognition models}.
\newblock In \emph{Neural Information Processing Systems}.

\bibitem[{Chen et~al.(2023)Chen, Hu, Luan, Sun, Changpinyo, Ritter, and
  Chang}]{Chen2023CanPV}
Yang Chen, Hexiang Hu, Yi~Luan, Haitian Sun, Soravit Changpinyo, Alan Ritter,
  and Ming-Wei Chang. 2023.
\newblock Can pre-trained vision and language models answer visual
  information-seeking questions?
\newblock In \emph{EMNLP}.

\bibitem[{Chen et~al.(2022)Chen, Duan, Wang, He, Lu, Dai, and
  Qiao}]{chen2022vitadapter}
Zhe Chen, Yuchen Duan, Wenhai Wang, Junjun He, Tong Lu, Jifeng Dai, and
  Yu~Qiao. 2022.
\newblock Vision transformer adapter for dense predictions.
\newblock \emph{arXiv preprint arXiv:2205.08534}.

\bibitem[{Finn et~al.(2017)Finn, Abbeel, and Levine}]{Finn2017ModelAgnosticMF}
Chelsea Finn, P.~Abbeel, and Sergey Levine. 2017.
\newblock Model-agnostic meta-learning for fast adaptation of deep networks.
\newblock \emph{ArXiv}, abs/1703.03400.

\bibitem[{Hu et~al.(2023)Hu, Luan, Chen, Khandelwal, Joshi, Lee, Toutanova, and
  Chang}]{Hu2023OpendomainVE}
Hexiang Hu, Yi~Luan, Yang Chen, Urvashi Khandelwal, Mandar Joshi, Kenton Lee,
  Kristina Toutanova, and Ming-Wei Chang. 2023.
\newblock Open-domain visual entity recognition: Towards recognizing millions
  of wikipedia entities.
\newblock \emph{ArXiv}, abs/2302.11154.

\bibitem[{Jia et~al.(2021)Jia, Yang, Xia, Chen, Parekh, Pham, Le, Sung, Li, and
  Duerig}]{Jia2021ScalingUV}
Chao Jia, Yinfei Yang, Ye~Xia, Yi-Ting Chen, Zarana Parekh, Hieu Pham, Quoc~V.
  Le, Yun-Hsuan Sung, Zhen Li, and Tom Duerig. 2021.
\newblock \href {https://api.semanticscholar.org/CorpusID:231879586} {Scaling
  up visual and vision-language representation learning with noisy text
  supervision}.
\newblock In \emph{International Conference on Machine Learning}.

\bibitem[{Johnson et~al.(2015)Johnson, Karpathy, and
  Fei-Fei}]{Johnson2015DenseCapFC}
Justin Johnson, Andrej Karpathy, and Li~Fei-Fei. 2015.
\newblock \href {https://api.semanticscholar.org/CorpusID:14521054} {Densecap:
  Fully convolutional localization networks for dense captioning}.
\newblock \emph{2016 IEEE Conference on Computer Vision and Pattern Recognition
  (CVPR)}, pages 4565--4574.

\bibitem[{Johnson et~al.(2016)Johnson, Karpathy, and Fei-Fei}]{densecap}
Justin Johnson, Andrej Karpathy, and Li~Fei-Fei. 2016.
\newblock Densecap: Fully convolutional localization networks for dense
  captioning.
\newblock In \emph{Proceedings of the IEEE Conference on Computer Vision and
  Pattern Recognition}.

\bibitem[{Kirillov et~al.(2023)Kirillov, Mintun, Ravi, Mao, Rolland, Gustafson,
  Xiao, Whitehead, Berg, Lo, Doll{\'a}r, and Girshick}]{Kirillov2023SegmentA}
Alexander Kirillov, Eric Mintun, Nikhila Ravi, Hanzi Mao, Chloe Rolland, Laura
  Gustafson, Tete Xiao, Spencer Whitehead, Alexander~C. Berg, Wan-Yen Lo, Piotr
  Doll{\'a}r, and Ross~B. Girshick. 2023.
\newblock Segment anything.
\newblock \emph{ArXiv}, abs/2304.02643.

\bibitem[{Kuo et~al.(2022)Kuo, Cui, Gu, Piergiovanni, and
  Angelova}]{Kuo2022FVLMOO}
Weicheng Kuo, Yin Cui, Xiuye Gu, A.~J. Piergiovanni, and Anelia Angelova. 2022.
\newblock F-vlm: Open-vocabulary object detection upon frozen vision and
  language models.
\newblock \emph{ArXiv}, abs/2209.15639.

\bibitem[{Kuznetsova et~al.(2018)}]{Kuznetsova2018TheOI}
Alina Kuznetsova et~al. 2018.
\newblock The open images dataset v4.
\newblock \emph{International Journal of Computer Vision}, 128:1956--1981.

\bibitem[{Lampert et~al.(2014)Lampert, Nickisch, and
  Harmeling}]{Lampert2014AttributeBasedCF}
Christoph~H. Lampert, Hannes Nickisch, and Stefan Harmeling. 2014.
\newblock Attribute-based classification for zero-shot visual object
  categorization.
\newblock \emph{IEEE Transactions on Pattern Analysis and Machine
  Intelligence}, 36:453--465.

\bibitem[{Li et~al.(2022)Li, Li, Xiong, and Hoi}]{Li2022BLIPBL}
Junnan Li, Dongxu Li, Caiming Xiong, and Steven C.~H. Hoi. 2022.
\newblock Blip: Bootstrapping language-image pre-training for unified
  vision-language understanding and generation.
\newblock In \emph{International Conference on Machine Learning}.

\bibitem[{Li et~al.(2021)Li, Zhang, Zhang, Yang, Li, Zhong, Wang, Yuan, Zhang,
  Hwang, Chang, and Gao}]{Li2021GroundedLP}
Liunian~Harold Li, Pengchuan Zhang, Haotian Zhang, Jianwei Yang, Chunyuan Li,
  Yiwu Zhong, Lijuan Wang, Lu~Yuan, Lei Zhang, Jenq-Neng Hwang, Kai-Wei Chang,
  and Jianfeng Gao. 2021.
\newblock Grounded language-image pre-training.
\newblock \emph{2022 IEEE/CVF Conference on Computer Vision and Pattern
  Recognition (CVPR)}, pages 10955--10965.

\bibitem[{Lin(2004)}]{Lin2004ROUGEAP}
Chin-Yew Lin. 2004.
\newblock Rouge: A package for automatic evaluation of summaries.
\newblock In \emph{ACL 2004}.

\bibitem[{Lin et~al.(2014)Lin, Maire, Belongie, Hays, Perona, Ramanan,
  Doll{\'a}r, and Zitnick}]{Lin2014MicrosoftCC}
Tsung-Yi Lin, Michael Maire, Serge~J. Belongie, James Hays, Pietro Perona, Deva
  Ramanan, Piotr Doll{\'a}r, and C.~Lawrence Zitnick. 2014.
\newblock Microsoft coco: Common objects in context.
\newblock In \emph{ECCV}.

\bibitem[{Liu et~al.(2019)Liu, Miao, Zhan, Wang, Gong, and
  Yu}]{Liu2019LargeScaleLR}
Ziwei Liu, Zhongqi Miao, Xiaohang Zhan, Jiayun Wang, Boqing Gong, and Stella~X.
  Yu. 2019.
\newblock Large-scale long-tailed recognition in an open world.
\newblock \emph{2019 IEEE/CVF Conference on Computer Vision and Pattern
  Recognition (CVPR)}, pages 2532--2541.

\bibitem[{Nguyen et~al.(2022)Nguyen, Suganuma, and Okatani}]{Nguyen2022GRITFA}
Van-Quang Nguyen, Masanori Suganuma, and Takayuki Okatani. 2022.
\newblock Grit: Faster and better image captioning transformer using dual
  visual features.
\newblock \emph{ArXiv}, abs/2207.09666.

\bibitem[{OpenAI(2023)}]{OpenAI2023GPT4TR}
OpenAI. 2023.
\newblock \href {https://api.semanticscholar.org/CorpusID:257532815} {Gpt-4
  technical report}.
\newblock \emph{ArXiv}, abs/2303.08774.

\bibitem[{Ordonez et~al.(2011)Ordonez, Kulkarni, and Berg}]{NIPS2011_5dd9db5e}
Vicente Ordonez, Girish Kulkarni, and Tamara Berg. 2011.
\newblock Im2text: Describing images using 1 million captioned photographs.
\newblock In \emph{Advances in Neural Information Processing Systems},
  volume~24. Curran Associates, Inc.

\bibitem[{Papineni et~al.(2002)Papineni, Roukos, Ward, and
  Zhu}]{Papineni2002BleuAM}
Kishore Papineni, Salim Roukos, Todd Ward, and Wei-Jing Zhu. 2002.
\newblock Bleu: a method for automatic evaluation of machine translation.
\newblock In \emph{ACL}.

\bibitem[{Peng et~al.(2022)Peng, Dong, Bao, Ye, and Wei}]{Peng2022BEiTVM}
Zhiliang Peng, Li~Dong, Hangbo Bao, Qixiang Ye, and Furu Wei. 2022.
\newblock \href {https://api.semanticscholar.org/CorpusID:251554649} {Beit v2:
  Masked image modeling with vector-quantized visual tokenizers}.
\newblock \emph{ArXiv}, abs/2208.06366.

\bibitem[{Qiu et~al.(2024)Qiu, Madotto, Lin, Crook, Xu, Dong, Faloutsos, Li,
  Damavandi, and Moon}]{qiu2024snapntell}
Jielin Qiu, Andrea Madotto, Zhaojiang Lin, Paul~A Crook, Yifan~Ethan Xu,
  Xin~Luna Dong, Christos Faloutsos, Lei Li, Babak Damavandi, and Seungwhan
  Moon. 2024.
\newblock Snapntell: Enhancing entity-centric visual question answering with
  retrieval augmented multimodal llm.
\newblock \emph{arXiv preprint arXiv:2403.04735}.

\bibitem[{Radford et~al.(2021)Radford, Kim, Hallacy, Ramesh, Goh, Agarwal,
  Sastry, Askell, Mishkin, Clark, Krueger, and
  Sutskever}]{Radford2021LearningTV}
Alec Radford, Jong~Wook Kim, Chris Hallacy, Aditya Ramesh, Gabriel Goh,
  Sandhini Agarwal, Girish Sastry, Amanda Askell, Pamela Mishkin, Jack Clark,
  Gretchen Krueger, and Ilya Sutskever. 2021.
\newblock \href {https://api.semanticscholar.org/CorpusID:231591445} {Learning
  transferable visual models from natural language supervision}.
\newblock In \emph{International Conference on Machine Learning}.

\bibitem[{Redmon et~al.(2015)Redmon, Divvala, Girshick, and
  Farhadi}]{Redmon2015YouOL}
Joseph Redmon, Santosh~Kumar Divvala, Ross~B. Girshick, and Ali Farhadi. 2015.
\newblock You only look once: Unified, real-time object detection.
\newblock \emph{2016 IEEE Conference on Computer Vision and Pattern Recognition
  (CVPR)}, pages 779--788.

\bibitem[{Ren et~al.(2015)Ren, He, Girshick, and Sun}]{Ren2015FasterRT}
Shaoqing Ren, Kaiming He, Ross~B. Girshick, and Jian Sun. 2015.
\newblock Faster r-cnn: Towards real-time object detection with region proposal
  networks.
\newblock \emph{IEEE Transactions on Pattern Analysis and Machine
  Intelligence}, 39:1137--1149.

\bibitem[{Russakovsky et~al.(2014)Russakovsky, Deng, Su, Krause, Satheesh, Ma,
  Huang, Karpathy, Khosla, Bernstein, Berg, and
  Fei-Fei}]{Russakovsky2014ImageNetLS}
Olga Russakovsky, Jia Deng, Hao Su, Jonathan Krause, Sanjeev Satheesh, Sean Ma,
  Zhiheng Huang, Andrej Karpathy, Aditya Khosla, Michael~S. Bernstein,
  Alexander~C. Berg, and Li~Fei-Fei. 2014.
\newblock \href {https://api.semanticscholar.org/CorpusID:2930547} {Imagenet
  large scale visual recognition challenge}.
\newblock \emph{International Journal of Computer Vision}, 115:211 -- 252.

\bibitem[{Rusu et~al.(2018)Rusu, Rao, Sygnowski, Vinyals, Pascanu, Osindero,
  and Hadsell}]{Rusu2018MetaLearningWL}
Andrei~A. Rusu, Dushyant Rao, Jakub Sygnowski, Oriol Vinyals, Razvan Pascanu,
  Simon Osindero, and Raia Hadsell. 2018.
\newblock Meta-learning with latent embedding optimization.
\newblock \emph{ArXiv}, abs/1807.05960.

\bibitem[{Shao et~al.(2019)Shao, Li, Zhang, Peng, Yu, Zhang, Li, and
  Sun}]{Shao2019Objects365AL}
Shuai Shao, Zeming Li, Tianyuan Zhang, Chao Peng, Gang Yu, Xiangyu Zhang, Jing
  Li, and Jian Sun. 2019.
\newblock \href {https://api.semanticscholar.org/CorpusID:207967883}
  {Objects365: A large-scale, high-quality dataset for object detection}.
\newblock \emph{2019 IEEE/CVF International Conference on Computer Vision
  (ICCV)}, pages 8429--8438.

\bibitem[{Sharma et~al.()Sharma, Ding, Goodman, and
  Soricut}]{sharma-etal-2018-conceptual}
Piyush Sharma, Nan Ding, Sebastian Goodman, and Radu Soricut.
\newblock Conceptual captions: A cleaned, hypernymed, image alt-text dataset
  for automatic image captioning.
\newblock In \emph{Proceedings of the 56th Annual Meeting of the Association
  for Computational Linguistics}, pages 2556--2565.

\bibitem[{Snell et~al.(2017)Snell, Swersky, and
  Zemel}]{Snell2017PrototypicalNF}
Jake Snell, Kevin Swersky, and Richard~S. Zemel. 2017.
\newblock Prototypical networks for few-shot learning.
\newblock In \emph{NIPS}.

\bibitem[{Tan and Le(2019)}]{Tan2019EfficientNetRM}
Mingxing Tan and Quoc~V. Le. 2019.
\newblock \href {https://api.semanticscholar.org/CorpusID:167217261}
  {Efficientnet: Rethinking model scaling for convolutional neural networks}.
\newblock \emph{ArXiv}, abs/1905.11946.

\bibitem[{Vinyals et~al.(2016)Vinyals, Blundell, Lillicrap, Kavukcuoglu, and
  Wierstra}]{Vinyals2016MatchingNF}
Oriol Vinyals, Charles Blundell, Timothy~P. Lillicrap, Koray Kavukcuoglu, and
  Daan Wierstra. 2016.
\newblock Matching networks for one shot learning.
\newblock \emph{ArXiv}, abs/1606.04080.

\bibitem[{Wang et~al.(2022{\natexlab{a}})Wang, Yang, Hu, Li, Lin, Gan, Liu,
  Liu, and Wang}]{Wang2022GITAG}
Jianfeng Wang, Zhengyuan Yang, Xiaowei Hu, Linjie Li, Kevin Lin, Zhe Gan,
  Zicheng Liu, Ce~Liu, and Lijuan Wang. 2022{\natexlab{a}}.
\newblock Git: A generative image-to-text transformer for vision and language.
\newblock \emph{ArXiv}, abs/2205.14100.

\bibitem[{Wang et~al.(2022{\natexlab{b}})}]{Wang2022UnifyingAT}
Peng Wang et~al. 2022{\natexlab{b}}.
\newblock Unifying architectures, tasks, and modalities through a simple
  sequence-to-sequence learning framework.
\newblock In \emph{International Conference on Machine Learning}.

\bibitem[{Wu et~al.(2022)Wu, Wang, Yang, Gan, Liu, Yuan, and
  Wang}]{Wu2022GRiTAG}
Jialian Wu, Jianfeng Wang, Zhengyuan Yang, Zhe Gan, Zicheng Liu, Junsong Yuan,
  and Lijuan Wang. 2022.
\newblock \href {https://api.semanticscholar.org/CorpusID:254125445} {Grit: A
  generative region-to-text transformer for object understanding}.
\newblock \emph{ArXiv}, abs/2212.00280.

\bibitem[{Xiao et~al.(2010)}]{Xiao2010SUNDL}
Jianxiong Xiao et~al. 2010.
\newblock \href {https://api.semanticscholar.org/CorpusID:1309931} {Sun
  database: Large-scale scene recognition from abbey to zoo}.
\newblock \emph{2010 IEEE Computer Society Conference on Computer Vision and
  Pattern Recognition}, pages 3485--3492.

\bibitem[{Ye et~al.(2018)Ye, Hu, chuan Zhan, and Sha}]{Ye2018FewShotLV}
Han-Jia Ye, Hexiang Hu, De~chuan Zhan, and Fei Sha. 2018.
\newblock Few-shot learning via embedding adaptation with set-to-set functions.
\newblock \emph{2020 IEEE/CVF Conference on Computer Vision and Pattern
  Recognition (CVPR)}, pages 8805--8814.

\bibitem[{Zhang et~al.(2022{\natexlab{a}})Zhang, Li, Liu, Zhang, Su, Zhu, Ni,
  and Shum}]{zhang2022dino}
Hao Zhang, Feng Li, Shilong Liu, Lei Zhang, Hang Su, Jun Zhu, Lionel~M. Ni, and
  Heung-Yeung Shum. 2022{\natexlab{a}}.
\newblock \href {http://arxiv.org/abs/2203.03605} {Dino: Detr with improved
  denoising anchor boxes for end-to-end object detection}.

\bibitem[{Zhang et~al.(2022{\natexlab{b}})Zhang, Zhang, Hu, Chen, Li, Dai,
  Wang, Yuan, Hwang, and Gao}]{Zhang2022GLIPv2UL}
Haotian Zhang, Pengchuan Zhang, Xiaowei Hu, Yen-Chun Chen, Liunian~Harold Li,
  Xiyang Dai, Lijuan Wang, Lu~Yuan, Jenq-Neng Hwang, and Jianfeng Gao.
  2022{\natexlab{b}}.
\newblock Glipv2: Unifying localization and vision-language understanding.
\newblock \emph{ArXiv}, abs/2206.05836.

\bibitem[{Zhang et~al.(2020)Zhang, Kishore, Wu, Weinberger, and
  Artzi}]{Zhang2020BERTScoreET}
Tianyi Zhang, Varsha Kishore, Felix Wu, Kilian~Q. Weinberger, and Yoav Artzi.
  2020.
\newblock Bertscore: Evaluating text generation with bert.
\newblock \emph{ArXiv}, abs/1904.09675.

\end{thebibliography}
\bibliographystyle{acl_natbib}

\clearpage
\appendix
\section{More Details about Baselines}

\subsubsection{Object Detection}

\paragraph{GLIP}
For GLIP~\citep{Li2021GroundedLP}, we use the GLIP-T model that uses the Tiny Swin-Tiny backbone and pretrained on Object365 \citep{Shao2019Objects365AL}, GoldG \citep{Li2021GroundedLP}, Cap4M \citep{Li2021GroundedLP}, SBU \citep{NIPS2011_5dd9db5e}, and Conceptual Captions \citep{sharma-etal-2018-conceptual}. The backbone for the text encoder is the base BERT model. 

\paragraph{GRiT} 
For GRiT~\citep{Wu2022GRiTAG}, we use the base GRiT model pretrained with the 12-layer ViT initialized from the masked autoencoder (MAE), which was trained on ImageNet-1K.
The text decoder is a 6-layer transformer. The provided checkpoint is also pretrained jointly on object detection and dense captioning.

\paragraph{DINO}
For DINO~\citep{zhang2022dino}, we use the 24 epoch setting, DINO-4scale pretrained checkpoint. 
This pretrained model uses the ResNet50 as the backbone, where a 6-layer encoder and 6-layer decoder are used for the transformer network \citep{zhang2022dino}. 
The hidden dimension size is 256.

\paragraph{ViT-Adapter}
For ViT-Adapter~\citep{chen2022vitadapter}, we use the large model. The ViT has 24 layers with 16 heads and 303.3 million parameters. The adapter has 16 heads as well and 23.7 million parameters. The backbone used in this pretrained model is the BEiTv2 model \citep{Peng2022BEiTVM}. 

\subsubsection{Zero-shot Image Classification}

\paragraph{CLIP-ViT-L}
The CLIP \citep{Radford2021LearningTV} model we utilize uses the large ViT transformer architecture as the image encoder and a masked self-attention transformer as the text encoder. We used clip-vit-large-patch14 in this setting.

\paragraph{CLIP-ViT-H} 
This CLIP \citep{Radford2021LearningTV} rendition uses the huge ViT as the backbone and was trained on the English subset of LAION-5B. We used CLIP-ViT-H-14-laion2B-s32B-b79K in this setting.

\paragraph{ALIGN} 
The ALIGN model \citep{Jia2021ScalingUV} uses the EfficientNet \citep{Tan2019EfficientNetRM} as the vision encoder and the BERT model as the text encoder. We used ALIGN-base in this setting.

\paragraph{GPT4}

GPT-4 \citep{OpenAI2023GPT4TR} is a large multimodal model capable of processing image and text inputs and producing text outputs.

\subsubsection{Image Captioning}

\paragraph{BLIP}
For BLIP~\citep{Li2022BLIPBL}, we use the ``blip-image-captioning-large" pretrained checkpoint, where ViT-Large is used as the vision transformer and the Bert-base model for the text transformer \citep{Li2022BLIPBL}. 
We use the phrase ``a picture of" as the prompt for the model, as seen in \cite{Li2022BLIPBL}.

\paragraph{OFA} 
For OFA~\citep{Wang2022UnifyingAT}, we use the ``OFA-base" pretrained checkpoint, where ResNet101 is used as the backbone~\citep{Wang2022UnifyingAT}.
This model has 180 million parameters, a hidden size of 768, and an intermediate size of 3072. There are 12 heads, six encoder layers, and six decoder layers.

\vspace{-5pt}
\paragraph{GIT}  
For GIT~\citep{Wang2022GITAG}, we use the ``git-base-coco" pretrained checkpoint, which contains six layers for the transformer decoder with 12 attention heads. 
The hidden size is 768, and the model has 347 million parameters. 

\vspace{-5pt}
\paragraph{GRIT} 
For GRIT~\citep{Nguyen2022GRITFA}, we use the checkpoint pretrained on four object detection datasets (i.e., COCO, Visual Genome, Open Images, and Object365) \citep{Nguyen2022GRITFA}.
The hidden size is set to 512, and the number of heads to 8. 
The model has six layers for the object detector, three layers for the grid feature network, and three layers for the caption generator \citep{Nguyen2022GRITFA}.

\subsubsection{Dense Captioning }

\paragraph{FCLN \citep{Johnson2015DenseCapFC}} 
FCLN uses a 13-layer VGG-16 architecture as the backbone and an RNN language model as the text decoder \citep{Johnson2015DenseCapFC}. The token and hidden layer size are 512.

\paragraph{GRiT-{MAE} \citep{Wu2022GRiTAG}} 
Similar to object detection, we use the base GRiT model pretrained with the 12-layer ViT initialized from the masked autoencoder (MAE).
The text decoder is also a 6-layer transformer. Since the provided checkpoint is jointly pretrained on object detection and dense captioning, we use the same checkpoint for the two tasks.

\begin{figure*}[tp]
  \centering
%  \fbox{\rule{0pt}{2in} \rule{0.9\linewidth}{0pt}}
  \includegraphics[width=0.7\linewidth]{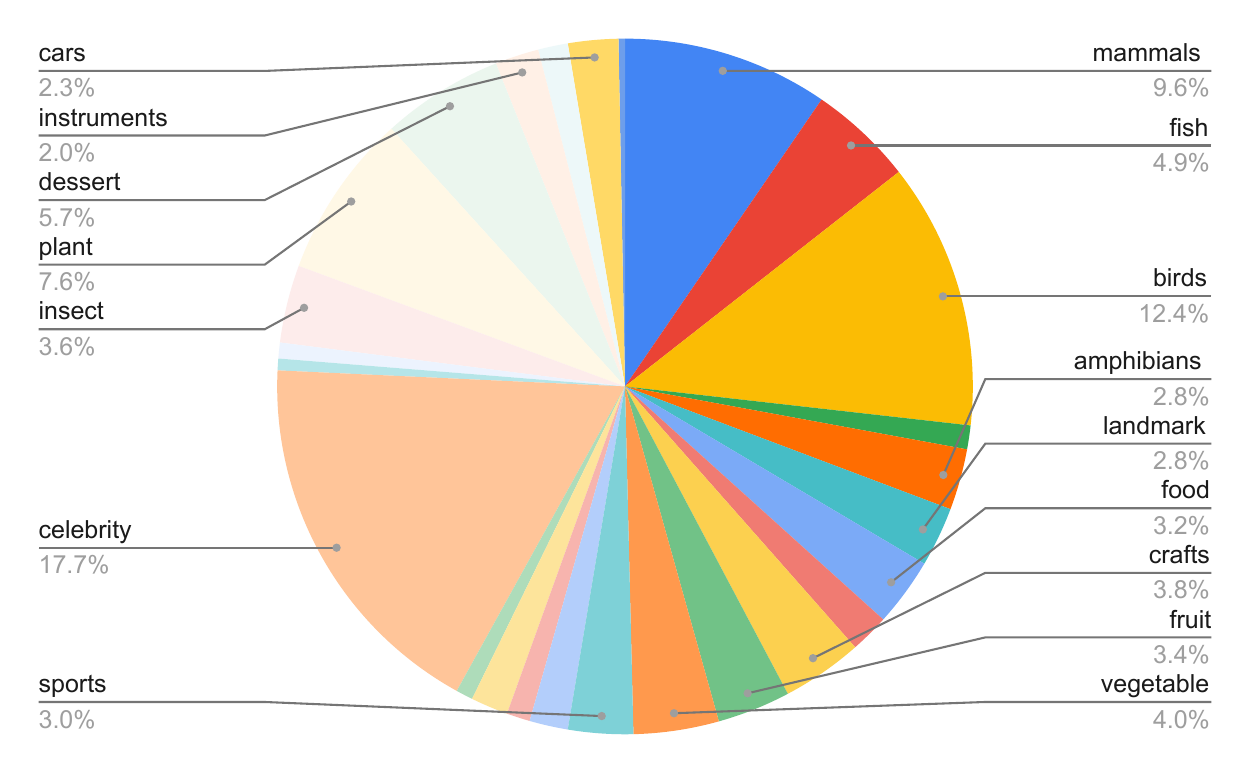}
  \caption{Statistics of the entities in each category.}
  \label{Fig:statistics}
  \vspace{-10pt}
\end{figure*}

\section{More Details about the Entity6K Dataset}\label{sec:appendix-dataset-details}

\subsection{Data Acquisition}

\paragraph{Entity List} 

Our initial step in addressing our problem involves the compilation of a diverse array of entity names, encompassing a wide range of real-world entities, including businesses, products, and individuals. To accomplish this task, we've categorized our selection into 26 distinct categories. Within each of these categories, we employed Wikipedia as a valuable resource to identify specific entity names. Our primary objective is to evaluate the system's capacity to accurately recognize precise entities, so we prioritize names that exhibit a high level of specificity. For instance, we favor names like ``German Shepherd" or ``Alaskan Malamute" over more general terms such as ``Dog." This unique approach differentiates our dataset from existing ones.

\vspace{-5pt}
\paragraph{Data Collection and Licenses}

After compiling a thorough and varied list of unique entities, ensuring there are no repetitions, the next step involves acquiring images. We accomplish this by utilizing the entity names as search queries on Flickr\footnote{https://www.flickr.com/photos/tags/dataset/}. It's important to note that these images have been generously shared on Flickr by their respective creators under licenses that include Creative Commons BY 2.0, Creative Commons BY-NC 2.0, Public Domain Mark, or Public Domain CC 1.0. These licenses all grant permission for unrestricted usage, redistribution, and modification, specifically for non-commercial purposes.

\vspace{-5pt}
\paragraph{Fidelity Control}

The dataset comprises 28,500 high-quality images with significant diversity, all sourced from Flickr, thereby inheriting the biases in that database.
Initially, we compiled 12,003 entity names across 26 categories. For each entity, we collected ten images from Flickr with approved licenses, saving the relevant metadata in a JSON file, including original image URLs, authors, and licenses. Subsequently, Amazon Mechanical Turk\footnote{https://www.mturk.com/} was employed to assess image quality through two key steps: (1) Three human judges verified if the saved image accurately corresponded to the entity; any mismatches led to image deletion. (2) Following this verification, entities lacking five saved images were removed from our list. For entities with more than five images, five were randomly sampled, forming our final dataset. After these fidelity control measures, we retained 5,700 entities, resulting in a retention rate of approximately 47.5\%. The detailed numbers of entities of each category before and after the fidelity control step are shown in Table~\ref{table:fidelity_control}.

\begin{figure*}[t]
  \centering
%  \fbox{\rule{0pt}{2in} \rule{0.9\linewidth}{0pt}}
  \includegraphics[width=0.99\linewidth]{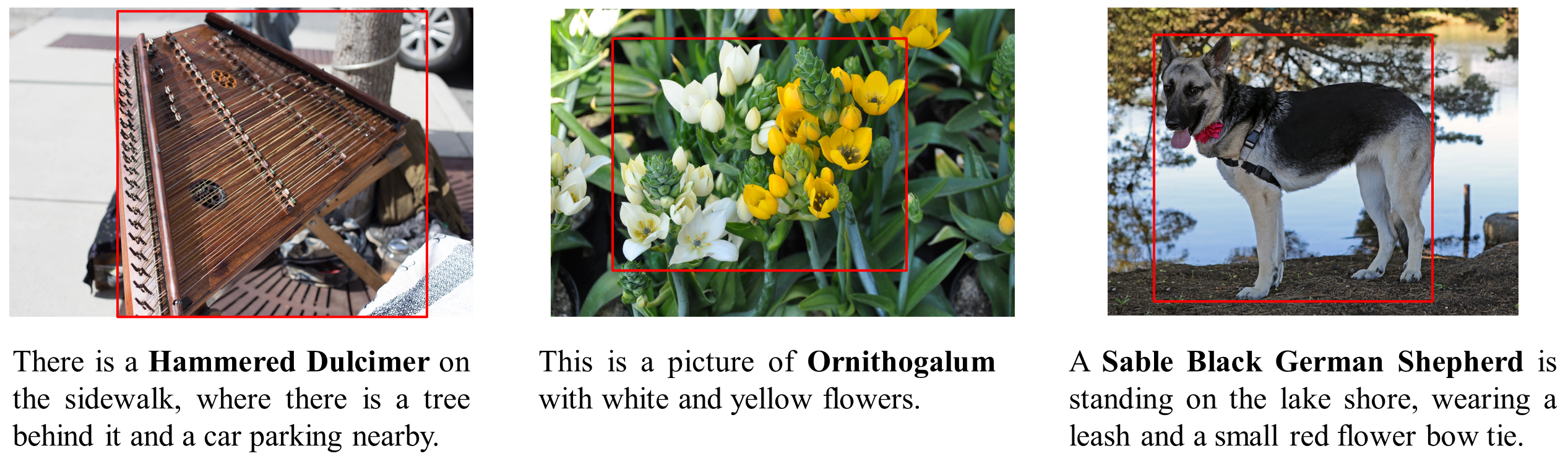}
  \vspace{-10pt}
  \caption{Examples of the collected data in the Entity6K dataset, where each image is associated with the entity region (bounding box) and the textual descriptions, \underline{centering on the specific entity}. }
  \label{Fig:example-appendix}
  \vspace{-10pt}
\end{figure*}

%\begin{figure*}[tp]
%  \centering
%  \fbox{\rule{0pt}{2in} \rule{0.9\linewidth}{0pt}}
%  \includegraphics[width=0.8\linewidth]{figs/pipeline.png}
%  \caption{The pipeline of the dataset building.}
%  \label{Fig:pipeline}
%\end{figure*}

\vspace{-5pt}
\subsection{Human Annotation}

The dataset labeling process comprises two distinct stages with Amazon Mechanical Turk:

\paragraph{Bounding Box Annotation} In the initial phase, a single annotator is assigned the task of outlining bounding boxes for each image. The annotator is provided with the corresponding entity name for the image and is responsible for marking the relevant region within that image. The goal is to establish a single bounding box for each image.

\paragraph{Textual Description Annotation} Following the completion of the initial bounding box marking phase by the first annotator, the second step involves five different annotators independently creating textual descriptions for each image. These annotators are given the entity name associated with each image to assist them in crafting their text captions. It's crucial to emphasize that all annotators are expected to provide comprehensive and detailed textual descriptions, encompassing as much relevant information as possible. For example, annotators are encouraged to write descriptions such as ``A cheerful boy, wearing a white helmet, is riding a vibrant green bicycle, while nearby, a young girl in a pink helmet is seated on a serene blue bicycle, sipping refreshing water" rather than simply stating ``Two people riding bikes."

\subsection{Statistics of the Dataset}

In Figure~\ref{Fig:statistics}, we can observe the statistics of the gathered Entity6K dataset. Furthermore, Table~\ref{table:compare_dataset} presents a comparison with existing datasets. As depicted in Table~\ref{table:compare_dataset}, our dataset contains an order of magnitude more entities than the existing datasets. Additionally, the entities are categorized and come with verified human annotations, rendering the proposed dataset a valuable resource for real-world entity recognition evaluations.

\begin{table}[tp]
\caption{More details for fidelity control, where ``Initial Entities" and ``Final Entities" mean the number of entities before/after the fidelity control step, respectively. }
\centering
\begin{adjustbox}{width=0.9\linewidth}
\begin{tabular}{l|r|r}\toprule
Main category &Initial Entities &Final Entities \\
\midrule
mammals &778 &545 \\
fish &1089 &277 \\
birds &739 &705 \\
reptiles &141 &63 \\
amphibians &211 &162 \\
landmark &500 &158 \\
food &483 &181 \\
electronics &432 &103 \\
crafts &490 &214 \\
fruit &361 &194 \\
vegetable &389 &226 \\
sports &694 &172 \\
household &120 &102 \\
games &198 &62 \\
toys &231 &99 \\
currency &157 &45 \\
celebrity &1515 &1009 \\
drink &300 &31 \\
healthcare &100 &42 \\
insect &369 &206 \\
plant &606 &436 \\
dessert &400 &323 \\
instruments &477 &116 \\
rock &217 &79 \\
cars &588 &133 \\
beauty &418 &17 \\
\midrule
Summary & 12,003 &5,700 \\
\bottomrule
\end{tabular}
\label{table:fidelity_control}
\end{adjustbox}
\vspace{-5pt}
\end{table}

\section{More Related Work}\label{sec:appendix-more-related-work}

\paragraph{Open-domain Entity Recognition} from images refers to the task of automatically identifying and extracting entities (objects, people, locations, etc.) from images without relying on any specific domain or prior knowledge.  There are few works in the open-domain entity recognition area. \citet{Hu2023OpendomainVE} presented the task of open-domain visual entity recognition, where a model needs to link an image to a Wikipedia entity with respect to a text query. However, their work needs a text query to retrieve the entity name in the Wikipedia entity name list. \citet{Chen2023CanPV} introduced INFOSEEK, a dataset for Visual Question Answering focused on informational queries. \citet{qiu2024snapntell} proposed a new task for entity-centric visual question answering to evaluate models' ability to understand identified entities.

\paragraph{Image Classification} aims at recognizing the class of the given image from a pre-defined class list. Recently, the zero-shot (ZS) setting has also been studied, where the classes are unseen in the training data \citep{Lampert2014AttributeBasedCF,Liu2019LargeScaleLR,Vinyals2016MatchingNF}. However, zero-shot seems to be a too complicated problem, and the few-shot setting has been considered, such as meta classifiers \citep{Snell2017PrototypicalNF,Finn2017ModelAgnosticMF,Rusu2018MetaLearningWL,Ye2018FewShotLV}.

\paragraph{Object Detection} algorithms, such as Faster R-CNN \citep{Ren2015FasterRT} or YOLO \citep{Redmon2015YouOL}, can be used to identify and localize objects within an image. These algorithms typically output bounding boxes around detected objects along with their corresponding class labels. \citet{Kuo2022FVLMOO} proposed F-VLM, an open-vocabulary object detection method built upon Frozen Vision and Language Models.  \citet{Li2021GroundedLP} proposed a grounded language-image pretraining (GLIP) model for learning object-level, language-aware, and semantic-rich visual representations, which unified object detection and phrase grounding for pre-training.  \citet{Zhang2022GLIPv2UL} unified localization pre-training and Vision-Language Pre-training, which can be used for object detection and instance segmentation.

\paragraph{Image Segmentation} techniques can be employed to partition an image into different regions or segments corresponding to different entities. This approach can provide more fine-grained entity recognition by precisely delineating the boundaries of objects in an image. \citet{Kirillov2023SegmentA} lifted image segmentation into the era of foundation models. The model is designed and trained to be promptable, so it can transfer zero-shot to new image distributions and tasks.

\section{Detailed Results for Each Category}\label{sec:appendix-more-results}

We also provided the detailed results for each category on each task. Tables~\ref{table:imgcap_results_ofa},\ref{table:imgcap_results_blip},\ref{table:imgcap_results_grit},\ref{table:imgcap_results_git} show detailed image captioning results by OFA, BLIP, GRiT, and GIT, respectively. Tables~\ref{table:obj_det_glip-cat},\ref{table:obj_det_grit-cat},\ref{table:obj_det_dino-cat},\ref{table:obj_det_vit-cat} show detailed object detection results for GLIP, GRiT, DINO, and ViT-Adapter, respectively. Table~\ref{table:zero_shot-cat} shows the detailed Zero-shot Image Classification results, and Table~\ref{table:dense_cap-cat} shows the detailed Dense Captioning across 26 categories.

\begin{table}[htp]\small
\caption{Comparison of accuracies for Zero-shot Image Classification across 26 categories. CLIP-ViT-L: CLIP-ViT-Large-patch14, CLIP-ViT-H: CLIP-ViT-H-14-laion2B-s32B-b79K.}
\centering
\begin{adjustbox}{width=0.95\linewidth}
\begin{tabular}{l|ccc}
\toprule
Category & CLIP-ViT-L   & CLIP-ViT-H  & ALIGN  \\
\midrule
crafts & 43.74 & 49.76 & 41.30 \\
mammals & 56.01 & 58.62 & 35.64 \\
food & 71.54 & 75.00 & 62.39 \\
plant & 50.51 & 52.16 & 22.87 \\
birds & 52.55 & 72.03 & 23.84 \\
fish & 31.07 & 37.50 & 16.55 \\
sports & 69.01 & 71.23 & 60.00 \\
dessert & 45.98 & 50.90 & 39.75 \\
celebrity & 80.86 & 71.92 & 38.32 \\
amphibians & 20.13 & 22.66 & 10.00 \\
vegetable & 42.21 & 43.63 & 31.08 \\
insect & 37.47 & 36.27 & 24.43 \\
healthcare & 49.76 & 54.63 & 56.10 \\
games & 58.00 & 62.00 & 40.67 \\
cars & 42.42 & 56.97 & 23.03 \\
fruit & 36.68 & 39.38 & 20.41 \\
electronics & 62.63 & 73.51 & 65.71 \\
toys & 35.32 & 40.76 & 38.68 \\ 
rock & 26.58 & 24.81 & 18.23 \\
household & 61.18 & 69.61 & 57.65 \\
instruments & 41.94 & 42.72 & 24.27 \\
landmark & \textbf{92.23} & \textbf{93.76} & 81.40 \\
reptiles & 41.90 & 42.22 & 23.17 \\
drink & 54.67 & 48.67 & 25.33 \\
currency & 65.78 & 62.22 & 36.44 \\
beauty & 81.18 & 92.94 & \textbf{89.41} \\
\bottomrule
\end{tabular}

\label{table:zero_shot-cat}
\end{adjustbox}
\vspace{-5pt}
\end{table}

\begin{table}[htp]\small
\caption{Comparison of mAP scores for Dense Captioning across 26 categories.}
\centering
\begin{adjustbox}{width=0.75\linewidth}
\begin{tabular}{l|cc}
\toprule
Category & FCLN & GRiT\_MAE \\
\midrule
crafts & 0.05 & 1.04 \\
mammals & 0.04 & 1.52 \\
food & 0.05 & 1.28 \\
plant & 0.02 & \textbf{3.04} \\
birds & 0.00 & 3.08 \\
fish & 0.01 & 1.76 \\
sports & 0.05 & 0.48 \\
dessert & 0.04 & 0.40 \\
celebrity & 0.01 & 2.64 \\
amphibians & 0.02 & 0.88 \\
vegetable & 0.00 & 3.00 \\
insect & 0.01 & 2.88 \\
healthcare & 0.02 & 1.92 \\
games & 0.04 & 0.64 \\
cars & 0.06 & 0.16 \\
fruit & 0.00 & 2.96 \\
electronics & 0.06 & 0.00 \\
toys & 0.03 & 1.50 \\ 
rock & 0.01 & 2.72 \\
household & 0.06 & 1.20 \\
instruments & 0.05 & 0.88 \\
landmark & 0.00 & 2.08 \\
reptiles & 0.06 & 0.32 \\
drink & 0.02 & 3.00 \\
currency & 0.00 & 2.80 \\
beauty & 0.04 & 1.52 \\
\bottomrule
\end{tabular}

\label{table:dense_cap-cat}
\end{adjustbox}
\vspace{-5pt}
\end{table}

\begin{table*}[htp]\small
\caption{Comparison of Image Captioning results for each category for OFA.}
\centering
\begin{adjustbox}{width=0.9\linewidth}
\begin{tabular}{l|rrrrrrr} 
\toprule
Category & ROUGE-1 $\uparrow$ & ROUGE-2 $\uparrow$ & ROUGE-L $\uparrow$ & BLEU $\uparrow$ & METEOR $\uparrow$ & SPICE $\uparrow$ & BertScore $\uparrow$  \\ 
\midrule
crafts & 12.14 & 1.12 & 10.79 & 0.91 & 6.16 & 1.34 & 84.61 \\
mammals & 9.93 & 1.18 & 9.07 & 0.44 & 4.79 & 1.34 & 83.94 \\
food & 20.20 & 3.72 & 17.34 & 2.02 & 11.05 &6.06& 85.52 \\
plant & 12.78 & 1.48 & 11.04 & 0.91 & 6.85 &5.05& 84.47 \\
birds & 11.98 & 2.43 & 10.87 & 0.47 & 5.23 & 1.82 & 84.04 \\
fish & 11.54 & 1.75 & 10.61 & 0.85 & 5.94 &1.19& 84.66 \\
sports & 16.74 & 3.04 & 14.29 & 0.80 & 7.93 & 5.42 & 85.68 \\
dessert & \textbf{20.21} & 3.64 & 17.34 & 1.58 & 10.64 & 7.40 & 85.91 \\
celebrity & 11.10 & 1.10 & 9.72 & 0.76 & 5.20 & 1.33 & 84.10 \\
amphibians & 12.71 & 2.44 & 11.55 & 0.99 & 7.46 & 3.00 & 85.35 \\
vegetable & 13.15 & 1.69 & 11.76 & 1.16 & 6.76 & 3.32&85.12 \\
insect & 14.96 & 2.78 & 13.64 & 0.81 & 8.01 & 3.24 &85.39 \\
healthcare &11.85 & 0.94 & 10.50 & 0.98& 5.68 & 1.05&85.13 \\
games & 17.31 & 2.22 & 14.63 & 0.96 & 7.85 & 6.42&85.35 \\
cars & 15.38 & \textbf{5.37} & 13.58 & 0.76 & 6.95 & 4.56 & 84.93 \\
fruit & 17.06 & 3.14 & 15.13 & 1.52 & 9.18 & 4.91 & 85.40 \\
electronics & 16.30 & 2.44 & 14.28 & 1.35 & 8.67 &5.82 & 86.14 \\
toys & 17.32 & 3.37 & 14.62 & 1.56 & 9.43 & 6.18& 85.67 \\
rock & 14.64 & 1.65 & 12.93 & 1.23 & 7.97 & 2.17 & 85.19 \\
household & 21.86 & 4.32 & \textbf{19.48} & \textbf{2.38} & \textbf{12.37} & \textbf{9.08} & \textbf{86.85} \\
instruments & 14.13 & 1.83 & 12.24 & 1.24 & 7.64 & 3.23& 84.76 \\
landmark & 12.96 & 1.74 & 11.05 & 0.73 & 7.87 & 6.64& 83.86 \\
reptiles & 11.63 & 2.14 & 10.67 & 0.65 & 5.96 &1.03 & 84.43 \\
drink & 17.72 & 1.95 & 15.29 & 1.14 & 8.55 & 4.50& 84.98 \\
currency & 18.47 & 3.71 & 15.65 & 1.44 & 8.05 & 4.37& 84.34 \\
beauty & 14.04 & 2.26 & 12.66 & 1.11 & 7.72 & 1.88& 84.89 \\
\bottomrule
\end{tabular}

\label{table:imgcap_results_ofa}
\end{adjustbox}
\end{table*}

\begin{table*}[htp]\small
\caption{Comparison of Image Captioning results for each category for BLIP.}
\centering
\begin{adjustbox}{width=0.9\linewidth}
\begin{tabular}{l|rrrrrrr} 
\toprule
Category & ROUGE-1 $\uparrow$ & ROUGE-2 $\uparrow$ & ROUGE-L $\uparrow$ & BLEU $\uparrow$ & METEOR $\uparrow$ & SPICE $\uparrow$ & BertScore $\uparrow$  \\
\midrule
crafts & 15.67 & 1.20 & 12.98 & 1.34 & 9.40  & 3.79 & 85.46 \\
mammals & 10.58 & 0.23 & 9.28 & 0.76 & 5.44 & 0.60 & 84.62 \\
food & 15.64 & 0.19 & 12.58 & 1.37 & 8.49  & 1.64 & 83.95 \\
plant & 9.06 & 0.20 & 8.33 & 0.93 & 4.52  & 0.49 & 83.24 \\
birds & 11.84 & 0.33 & 10.50 & 0.98 & 7.11  & 0.31 & 84.10 \\
fish & 14.07 & 0.16 & 12.62 & 1.15 & 7.60  & 0.96 & 84.53 \\
sports & \textbf{18.91} & 1.53 & 14.67 & 0.97 & 10.27& 5.42 & \textbf{86.30} \\
dessert & 14.40 & 0.27 & 11.87 & 1.08 & 7.84& 1.51 & 84.17 \\
celebrity & 18.68 & 1.99 & 14.90 & 1.36 & 10.41& 3.60 & 84.95 \\
amphibians & 10.95 & 0.39 & 10.13 & 1.16 & 7.67 & 0.67 & 85.09 \\
vegetable & 11.06 & 0.23 & 9.71 & 1.09 & 5.98& 0.31 & 84.69 \\
insect & 10.91 & 0.61 & 9.98 & 1.00 & 6.47& 0.37 & 84.34 \\
healthcare & 14.54 & 0.48 & 12.01 & 1.22 & 7.27 & 3.41 & 85.56 \\
games & 13.11 & 0.36 & 10.80 & 0.91 & 6.16  & \textbf{5.93} & 85.47 \\
cars & 14.16 & 1.49 & 10.51 & 0.74 & 7.48 & 0.75 & 84.13 \\
fruit & 11.99 & 0.59 & 10.60 & 1.24 & 6.97 & 0.42 & 84.32 \\
electronics & 13.18 & 0.38 & 11.61 & 1.22 & 7.97 & 0.93 & 85.34 \\
toys & 14.73 & 1.21 & 12.47 & 1.27 & 8.83  & 2.18 & 85.47 \\
rock & 13.16 & 0.11 & 11.48 & 1.20 & 6.75 & 3.44 & 84.44 \\
household & 13.70 & 0.67 & 11.66 & 1.46 & 8.36  & 1.12 & 85.47 \\
instruments & 16.10 & 1.77 & 13.25 & 1.57 & \textbf{10.47}  & 3.22 & 85.32 \\
landmark & 13.92 & 0.79 & 11.52 & 0.95 & 6.20 & 1.21 & 84.36 \\
reptiles & 10.33 & 0.40 & 9.27 & 0.99 & 6.97  & 0.63 & 84.52 \\
drink & 17.81 & 0.14 & 13.19 & 1.13 & 9.24  & 2.07 & 84.81 \\
currency & 18.27 & \textbf{4.61} & \textbf{15.38} & \textbf{1.91} & 10.36  & 4.65 & 84.84 \\
beauty & 13.46 & 0.89 & 10.55 & 1.14 & 8.48  & 1.12 & 84.71 \\
\bottomrule
\end{tabular}

\label{table:imgcap_results_blip}
\end{adjustbox}
\end{table*}

\begin{table*}[htp]\small
\caption{Comparison of Image Captioning results for each category for GRiT.}
\centering
\begin{adjustbox}{width=0.9\linewidth}
\begin{tabular}{l|rrrrrrr} 
\toprule
Category & ROUGE-1 $\uparrow$ & ROUGE-2 $\uparrow$ & ROUGE-L $\uparrow$ & BLEU $\uparrow$ & METEOR $\uparrow$ & SPICE $\uparrow$ & BertScore $\uparrow$  \\
\midrule
crafts&0.24&0.00&0.24&0.03&0.28&0.14&78.44 \\
mammals&0.05&0.00&0.05&0.00&0.19&0.11&77.77\\
food&0.26&0.00&0.26&0.03&0.25&0.05&77.80\\
plant&0.16&0.00&0.16&0.02&0.30&0.23&77.50 \\
birds&0.12&0.00&0.12&0.01&0.15&0.11&77.52 \\
fish&0.05&0.00&0.05&0.01&0.14&0.06&78.07 \\
sports&0.20&0.00&0.20&0.02&0.32&0.17&78.14\\
dessert&0.19&0.00&0.19&0.02&0.19&0.12&78.01\\
celebrity&0.08&0.00&0.08&0.01&0.17&0.13&77.66 \\
amphibians&0.04&0.00&0.04&0.01&0.13&0.07&78.45\\
vegetable&\textbf{0.29}&0.00&\textbf{0.29}&0.03&0.31&0.13&78.38\\
insect&0.06&0.00&0.05&0.01&0.14&0.08&77.84 \\
healthcare&0.16&0.00&0.16&0.02&0.24&0.12&78.52\\
games&0.16&0.00&0.16&0.02&0.37&0.17&78.50 \\
cars&0.09&0.00&0.09&0.01&0.16&0.04&77.40\\
fruit&0.18&0.00&0.18&0.02&0.23&0.14&77.98\\
electronics&0.09&0.00&0.09&0.01&0.21&0.12&78.60\\
toys&0.15&0.00&0.15&0.02&0.28&0.22&78.24\\
rock&0.05&0.00&0.05&0.01&0.19&0.10&78.14\\
household&0.20&0.00&0.20&0.02&0.21&0.21&\textbf{78.69} \\
instruments&0.03&0.00&0.03&0.00&0.13&0.08&78.28\\
landmark&0.09&0.00&0.09&0.01&0.12&0.17&78.05\\
reptiles&0.00&0.00&0.00&0.00&0.10&0.11&77.88\\
drink&0.15&0.00&0.15&0.02&0.28&0.04&78.33 \\
currency&0.27&0.00&0.27&0.03&\textbf{0.34}&\textbf{0.35}&77.60\\
beauty&0.16&0.00&0.16&0.02&0.18&0.34&78.20\\
\bottomrule
\end{tabular}

\label{table:imgcap_results_grit}
\end{adjustbox}
\end{table*}

\begin{table*}[htp]\small
\caption{Comparison of Image Captioning results for each category for GIT.}
\centering
\begin{adjustbox}{width=0.9\linewidth}
\begin{tabular}{l|rrrrrrr} 
\toprule
Category & ROUGE-1 $\uparrow$ & ROUGE-2 $\uparrow$ & ROUGE-L $\uparrow$ & BLEU $\uparrow$ & METEOR $\uparrow$ & SPICE $\uparrow$ & BertScore $\uparrow$  \\
\midrule
crafts & 12.89 & 1.29 & 11.34 & 0.59 & 5.01  & 2.15 & 82.43 \\
mammals & 9.51 & 0.84 & 8.10 & 0.17 & 3.91  & 1.95 & 80.68 \\
food & 13.79 & 0.59 & 12.19 & 0.81 & 5.33 & 2.16 & 80.87 \\
plant & 8.21 & 0.34 & 7.19 & 0.41 & 3.30  & 0.56 & 80.42 \\
birds & 12.70 & 1.16 & 11.19 & 0.30 & 4.59  & 0.75 & 80.76 \\
fish & 12.17 & 0.75 & 10.97 & 0.50 & 4.83  & 0.49 & 81.41 \\
sports & 9.87 & 0.82 & 8.77 & 0.13 & 3.16  & 1.06 & 81.58 \\
dessert & 11.41 & 0.50 & 10.32 & 0.39 & 4.11  & 1.56 & 81.17 \\
celebrity & 11.73 & 1.92 & 10.18 & 0.34 & 4.18 & 0.55 & 81.39 \\
amphibians & 10.17 & 0.32 & 8.30 & 0.49 & 5.44  & 2.57 & 81.49 \\
vegetable & 10.40 & 0.52 & 8.69 & 0.51 & 4.39  & 1.68 & 81.76 \\
insect & 12.07 & 1.08 & 10.19 & 0.37 & 4.87  & 0.58 & 81.27 \\
healthcare & 9.37 & 0.82 & 8.34 & 0.32 & 3.70  & 2.41 & 82.05 \\
games & 10.20 & 0.64 & 8.92 & 0.24 & 3.44  & 0.43 & 80.24 \\
cars & 11.42 & 2.30 & 10.23 & 0.09 & 4.04  & 0.60 & 80.92 \\
fruit & 11.96 & 0.93 & 10.76 & 0.67 & 5.03  & 0.98 & 81.57 \\
electronics & 12.79 & 1.80 & 10.97 & 0.68 & 5.12  & 0.69 & 82.84 \\
toys & 11.23 & 1.20 & 9.83 & 0.39 & 4.23  & 1.31 & 82.37 \\
rock & 12.93 & 0.64 & 11.28 & 0.61 & 5.04  & \textbf{4.44} & 82.41 \\
household & 12.23 & 0.98 & 10.66 & \textbf{0.77} & 4.94  & 1.26 & \textbf{82.99} \\
instruments & 11.98 & 1.35 & 10.65 & 0.53 & 4.67  & 0.62 & 82.11 \\
landmark & 9.92 & 0.59 & 8.70 & 0.23 & 3.28  & 0.70 & 81.20 \\
reptiles & 11.49 & 0.56 & 8.61 & 0.32 & \textbf{5.31}  & 2.06 & 80.87 \\
drink & 12.27 & 0.25 & 10.41 & 0.32 & 4.14  & 1.79 & 81.28 \\
currency & \textbf{14.77} & \textbf{3.41} & \textbf{13.52} & 0.40 & 5.20  & 0.36 & 81.78 \\
beauty & 12.67 & 1.56 & 10.70 & 0.47 & 5.02 & 0.97 & 82.39 \\
\bottomrule
\end{tabular}

\label{table:imgcap_results_git}
\end{adjustbox}
\end{table*}

\begin{table}[htp]\small
\caption{Comparison of Object Detection results for each category for GLIP.}
\centering
\begin{adjustbox}{width=0.75\linewidth}
\begin{tabular}{l|rrr}
\toprule
    Category & AP & AP50 & AP75 \\
\midrule
crafts      & 16.59 &  7.54 &  0.01 \\
mammals     &  9.61 &  1.69 &  0.05 \\
food        &  0.00 & 13.16 &  0.04 \\
plant       & 10.19 &  4.82 &  0.08 \\
birds       &  1.25 &  0.00 &  0.05 \\
fish        &  4.96 & 13.92 &  0.00 \\
sports      & 18.85 & 14.61 &  0.07 \\
dessert     &  0.00 & 20.39 &  0.06 \\
celebrity   &  3.30 & 11.43 &  0.03 \\
amphibians  & 11.62 & 29.76 &  0.03 \\
vegetable   & 17.19 & 25.86 &  0.00 \\
insect      &  0.00 & 21.98 &  0.08 \\
healthcare  &  0.00 & 30.20 &  0.05 \\
games       & \textbf{19.78} &  0.00 &  0.05 \\
cars        & 10.87 &  5.15 &  0.00 \\
fruit       &  9.25 &  4.44 &  0.01 \\
electronics & 19.65 & \textbf{32.99} &  0.04 \\
toys        & 10.44 &  0.00 &  0.00 \\
rock        &  5.41 & 19.56 &  0.06 \\
household   &  0.00 & 11.46 &  0.09 \\
instruments & 18.43 &  0.00 &  0.05 \\
landmark    &  1.75 & 14.77 &  0.08 \\
reptiles    &  5.91 &  0.00 &  0.07 \\
drink       & 17.51 &  8.95 &  0.00 \\
currency    & 15.93 &  9.82 &  0.00 \\
beauty      &  2.92 & 23.55 &  0.06 \\
\bottomrule
\end{tabular}

\label{table:obj_det_glip-cat}
\end{adjustbox}
\end{table}

\begin{table}[htp]\small
\caption{Comparison of Object Detection results for each category for GRiT.}
\centering
\begin{adjustbox}{width=0.75\linewidth}
\begin{tabular}{l|rrr}
    \toprule
    Category & AP & AP50 & AP75 \\
    \midrule
    crafts      & 7.85  & 16.18 & 3.01 \\
    mammals     & 5.93  & 13.90 & 1.43 \\
    food        & 13.96 & 25.67 & 6.25 \\
    plant       & 14.50 & 27.46 & 5.51 \\
    birds       & 2.75  & 6.92  & 0.66 \\
    fish        & 5.12  & 10.26 & 1.81 \\
    sports      & 20.69 & 33.80 & 10.76 \\
    dessert     & 36.19 & 49.91 & 23.93 \\
    celebrity   & 4.50  & 9.92  & 1.40 \\
    amphibians  & 5.13  & 11.27 & 0.51 \\
    vegetable   & 16.42 & 29.30 & 7.03 \\
    insect      & 5.16  & 12.12 & 1.19 \\
    healthcare  & 4.55  & 9.27  & 0.00 \\
    games       & 50.74 & 63.67 & 37.67 \\
    cars        & 16.95 & 31.36 & 6.06 \\
    fruit       & 25.77 & 37.41 & 16.99 \\
    electronics & 29.95 & 43.74 & 17.66 \\
    toys        & \textbf{51.89} & 62.86 & 38.16 \\
    rock        & 20.27 & 32.41 & 9.62 \\
    household   & 50.55 & \textbf{65.88} & 35.29 \\
    instruments & 41.67 & 55.34 & 29.51 \\
    landmark    & 44.16 & 58.09 & 29.94 \\
    reptiles    & 4.04  & 7.94  & 1.27 \\
    drink       & 4.40  & 11.33 & 2.00 \\
    currency    & 50.96 & 60.89 & \textbf{40.44} \\
    beauty      & 9.11  & 15.29 & 4.71 \\
    \bottomrule
\end{tabular}

\label{table:obj_det_grit-cat}
\end{adjustbox}
\end{table}

\begin{table}[htp]\small
\caption{Comparison of Object Detection results for each category for DINO.}
\centering
\begin{adjustbox}{width=0.75\linewidth}
\begin{tabular}{l|rrr}
\toprule
    Category & AP & AP50 & AP75 \\
\midrule
crafts      & 14.62 & 28.81 &  0.00 \\
mammals     &  8.45 & 34.14 &  4.39 \\
food        & 20.40 & 30.05 &  1.37 \\
plant       &  0.00 & 21.92 &  3.39 \\
birds       &  0.00 &  0.00 &  4.87 \\
fish        &  9.51 & 29.97 &  5.22 \\
sports      &  9.11 &  0.00 &  0.00 \\
dessert     & 15.18 & 12.08 &  0.13 \\
celebrity   & \textbf{24.92} & 19.48 &  1.39 \\
amphibians  & 20.22 & 21.81 &  0.58 \\
vegetable   &  4.74 & 18.28 &  0.92 \\
insect      &  9.40 &  0.00 &  1.79 \\
healthcare  &  0.00 &  4.53 &  0.00 \\
games       &  0.00 &  7.49 &  \textbf{5.89} \\
cars        &  8.43 & 17.09 &  4.87 \\
fruit       &  4.93 & \textbf{34.45} &  0.70 \\
electronics & 13.55 & 10.73 &  3.16 \\
toys        &  0.00 &  0.00 &  1.62 \\
rock        & 13.08 &  9.52 &  1.74 \\
household   & 18.71 &  5.45 &  5.81 \\
instruments & 22.93 & 11.39 &  3.64 \\
landmark    & 21.01 &  0.00 &  0.00 \\
reptiles    & 10.19 &  4.41 &  3.32 \\
drink       & 19.40 & 14.74 &  4.17 \\
currency    & 10.20 & 20.11 &  2.65 \\
beauty      &  2.33 & 18.46 &  0.00 \\
\bottomrule
\end{tabular}

\label{table:obj_det_dino-cat}
\end{adjustbox}
\end{table}

\begin{table}[htp]\small
\caption{Comparison of Object Detection results for each category for ViT-Adapter.}
\centering
\begin{adjustbox}{width=0.75\linewidth}
\begin{tabular}{l|rrr}
    \toprule
    Category & AP & AP50 & AP75 \\
    \midrule
    crafts      & 5.56  & 9.59  & 1.53 \\
    mammals     & 4.13  & 7.59  & 0.68 \\
    food        & 9.56  & 15.35 & 3.77 \\
    plant       & 9.58  & 16.22 & 2.94 \\
    birds       & 4.00  & 7.55  & 0.45 \\
    fish        & 5.24  & 9.66  & 0.82 \\
    sports      & 9.97  & 15.12 & 4.83 \\
    dessert     & 31.36 & 40.55 & 22.17 \\
    celebrity   & 2.03  & 3.58  & 0.48 \\
    amphibians  & 5.77  & 10.35 & 1.19 \\
    vegetable   & 10.52 & 17.40 & 3.65 \\
    insect      & 5.74  & 10.64 & 0.83 \\
    healthcare  & 4.37  & 6.73  & 2.01 \\
    games       & 27.11 & 32.62 & 21.60 \\
    cars        & 12.36 & 19.24 & 5.48 \\
    fruit       & 21.57 & 30.79 & 12.35 \\
    electronics & 29.44 & 38.82 & 20.06 \\
    toys        & 32.90 & 41.50 & 24.30 \\
    rock        & 15.33 & 23.62 & 7.05 \\
    household   & 36.25 & 44.57 & \textbf{27.92} \\
    instruments & 31.29 & 39.60 & 22.97 \\
    landmark    & 4.78  & 6.92  & 2.64 \\
    reptiles    & 2.63  & 4.76  & 0.49 \\
    drink       & 4.85  & 8.51  & 1.18 \\
    currency    & \textbf{41.35} & \textbf{46.53} & 36.17 \\
    beauty      & 3.95  & 6.02  & 1.87 \\
    \bottomrule
\end{tabular}

\label{table:obj_det_vit-cat}
\end{adjustbox}
\end{table}

%\section{Example Appendix}
%\label{sec:appendix}

%This is an appendix.

\end{document}